\documentclass[sigconf]{acmart}
\usepackage{subcaption}
\usepackage{times}
\usepackage{epsfig}
\usepackage{graphicx}
\usepackage{amsmath, xparse}
\usepackage{verbatim}

\usepackage{algorithm}
\usepackage{algpseudocode}
\usepackage{subcaption}
\usepackage{booktabs}
\newcommand*{\pcrFont}{\fontfamily{pcr}\selectfont}

\AtBeginDocument{%
  \providecommand\BibTeX{{%
    \normalfont B\kern-0.5em{\scshape i\kern-0.25em b}\kern-0.8em\TeX}}}

\setcopyright{none}
\renewcommand\footnotetextcopyrightpermission[1]{} % removes footnote with conference information in first column
\begin{document}

\title{Superevents: Towards Native Semantic Segmentation for Event-based Cameras}

\author{Weng Fei Low}
\authornote{Both authors contributed equally to this research.}
\affiliation{%
  \institution{National University of Singapore}
  \country{Singapore}
}

\author{Ankit Sonthalia}
\authornotemark[1]
\affiliation{%
  \institution{Birla Institute of Technology \&}
  \streetaddress{1 Th{\o}rv{\"a}ld Circle}
  \country{Science, Pilani, India}
}

\author{Zhi Gao}
\affiliation{%
  \institution{National University of Singapore}
  \country{Singapore}
}

\author{André van Schaik}
\affiliation{%
 \institution{Western Sydney University}
 \city{International Center for Neuromorphic Systems}
 \state{New South Wales}
 \country{Australia}}
\email{A.VanSchaik@westernsydney.edu.au}

\author{Bharath Ramesh}
\affiliation{%
 \institution{Western Sydney University}
 \city{International Center for Neuromorphic Systems}
 \state{New South Wales}
 \country{Australia}}
\email{B.Ramesh@westernsydney.edu.au}

\renewcommand{\shortauthors}{Fei and Sonthalia, et al.}

\begin{abstract}
Most successful computer vision models transform low-level features, such as Gabor filter responses, into richer representations of intermediate or mid-level complexity for downstream visual tasks. These mid-level representations have not been explored for event cameras, although it is especially relevant to the visually sparse and often disjoint spatial information in the event stream. By making use of locally consistent intermediate representations, termed as superevents, numerous visual tasks ranging from semantic segmentation, visual tracking, depth estimation shall benefit. In essence, superevents are perceptually consistent local units that delineate parts of an object in a scene. Inspired by recent deep learning architectures, we present a novel method that employs lifetime augmentation for obtaining an event stream representation that is fed to a fully convolutional network to extract superevents. Our qualitative and quantitative experimental results on several sequences of a benchmark dataset highlights the significant potential for event-based downstream applications.
\end{abstract}






\maketitle

\section{Introduction}
\label{sec:intro}
In the past decade, a novel class of visual sensors, known as Dynamic Vision Sensors (DVS), has attracted a lot of attention from the academia as well as the industry. This is fueled by its exceptional performance compared to conventional image sensors in terms of temporal resolution, latency, dynamic range and power consumption. Such breakthrough is achieved via a fundamentally different principle of operation. Conventional image sensors output a synchronous sequence of intensity frames, whereas the DVS outputs a sequence of events asynchronously, whereby each event indicates a change in intensity at a particular pixel location and time instant.
\par

\begin{figure}[t]
    \centering
        \begin{subfigure}{0.48\linewidth}
    		\centering
			\includegraphics[width=1\linewidth]{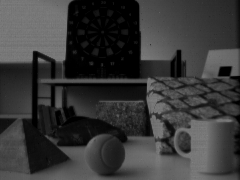} 
			\caption{}
			\label{bei:fig:slider_depth_aps}
		\end{subfigure}
		\begin{subfigure}{0.48\linewidth}
            \centering
            \includegraphics[width=1\linewidth]{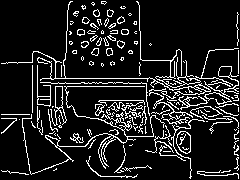}
            \caption{}
            \label{bei:fig:slider_depth_canny}
		\end{subfigure}
		\begin{subfigure}{0.48\linewidth}
            \centering
            \includegraphics[width=1\linewidth]{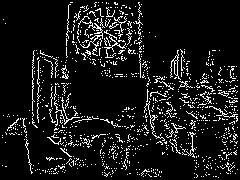}
            \caption{}
            \label{bei:fig:slider_depth_sofea_bei}
		\end{subfigure}
	    \begin{subfigure}{0.48\linewidth}
            \centering
            \includegraphics[width=1\linewidth]{{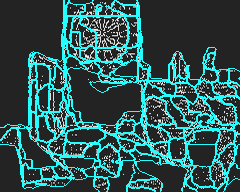}}
            \caption{}
            \label{bei:fig:slider_depth_mueggler2015}
	    \end{subfigure}
    \caption{(a). An event camera is moved on a slider in front of objects at different depth \cite{Mueggler2017}; (b) Edge map obtained via Canny edge detection is visualized for reference; (c) Scene-adaptive optical flow estimation is subsequently employed for lifetime estimation to produce crisp event stream images; (d) Superevents extracted using a fully convolutional network on the event stream image (best viewed on monitor).}
    \label{fig:intro_fig}
\end{figure}

Due to the above-mentioned unique advantages, event cameras have been trialed for a wide range of high-level vision tasks, namely object segmentation \cite{evseg2019}, motion segmentation \cite{Mitrokhin_2020_CVPR}, pattern recognition \cite{Ramesh2019, Ramesh2018}, gesture recognition \cite{eddd2020, chen2020}, space situational awareness \cite{bagchi2020event, chin2019star}, to name a few. Despite these advances in addressing these problems using conventional or deep learning methods that have made use of novel neural architecture design and training schemes suited for event-based vision, previous works have ignored mid-level representation that are critical in bridging gap between the sparse, disjoint information in the event stream and downstream visual tasks. As these novel cameras respond to structure and texture of the scene while not capturing intensity information, it is challenging to obtain groupings of events, i.e. mid-level representations, as done in superpixel extraction for image data \cite{stutz2018superpixels}. 
\par
In computer vision, superpixels are perceptually similar pixels grouped together to effectively reduce the number of image primitives for subsequent processing. This mid-level representation has been widely adopted for scene understanding\cite{wang2019revisiting} via detecting and segmenting salient regions \cite{zhao2019pyramid, zhao2015saliency, mostajabi2015feedforward}, and also for high-level tasks such as object detection \cite{li2018contour, shu2013improving} and tracking \cite{feng2019dynamic, yeo2017superpixel}. However, these ideas are yet to be explored in the event-based domain for mid-level feature extraction. One main reason is possibly that, the standard convolution operation in convolutional neural networks (CNNs) is defined on a regular image grid with intensity information. While a few attempts have been made to modify deep architectures \cite{asynconvet2018, asyncconvnet2020}, performing convolution over aysnchronous data remains challenging. 
\par
In this paper, we introduce a fully convolutional network (FCN) based superevents extraction algorithm that is in the spirit of \cite{yang2020sup}. The input to the FCN is a Binary Event Image (BEI) obtained via lifetime augmentation \cite{Mueggler2015}. In this regard, we make use of the open-source event-based flow estimation method SOFEA \cite{Low_2020_CVPR_Workshops}. It is worth noting that the recent works targeting super-resolution of the event stream\cite{mohd2020eventsr,wang2020eventsr} is very different to the task considered in this paper. Super-resolution aims to reconstruct high resolution representations directly from the event stream and is unrelated to superevents (or superpixels) mid-level feature extraction. 
\par
The primary contributions of this paper are:
\begin{enumerate}
\item A scene-adaptive noise rejection method for event-based optical flow estimation (\autoref{bei:subsec:application}) and a robust lifetime augmentation method for accurate BEI estimation (\autoref{bei:sec:lifetime}).
\item A scene-adaptive BEI rendering time interval which approximately normalizes for the expected apparent displacement magnitude of features with apparent motion (\autoref{bei:sec:render}).
\item A superevents extraction pipeline for event cameras leveraging on the above contributions using a fully convolutional network.
\end{enumerate}
%The rest of the paper is organized as follows: \autoref{sec:related} reviews the related work on estimating binary event representations and positions our work accordingly; \autoref{sec:bei} presents our SOFEA-BEI framework followed by the superevents extraction method in \autoref{sec:superevents}; \autoref{sec:experiments} contains the experimental results and discussion. The paper is concluded in \autoref{sec:conc}. 
%\begin{figure*}
%\begin{center}
%\fbox{\rule{0pt}{2in} \rule{.9\linewidth}{0pt}}
%\end{center}
%   \caption{Example of a short caption, which should be centered.}
%\label{fig:short}
%\end{figure*}

\section{Related Work}
\label{sec:related}
As previously mentioned, the output of a DVS is a stream of asynchronous events, as opposed to a set of images captured at a constant rate by a standard camera. Due to this unconventional output, the general design strategy of event-based algorithms has been to either implicitly or explicitly buffer a certain number of past events to perform a given task. The most straight-forward method is to accumulate events over a fixed time interval. For instance, \cite{Authors, Authorsbmvc19, Kogler2009} used contrived time intervals of $5-66 \mathit{ms}$ to estimate binary event images for object tracking and stereo vision. Nevertheless, the motion-dependent sensing aspect of the DVS is a major hindrance for choosing the optimal time interval to prevent motion blur or a low density event image. 
%Due to the fact that events are mainly generated along edges during motion \cite{Gallego2019}, a binary event image is visually similar to an edge map. However in reality, variation of illumination and noise exist, which manifest as artifacts in an event image. Moreover, motion may be along edges and hence no associated events would be generated.
\par
Another way to obtain an event stream representation is to implicitly choose a time interval by accumulating a fixed number of past events \cite{Ramesh2019, gallego2016event, Rebecq2016}. However, the critical issue is choosing the number of events, which unfortunately depends on the scene complexity and spatial resolution of the DVS itself. \cite{Liu2018} proposed a method to accumulate past events until the density of events in a local spatial neighborhood reaches a predefined threshold. It adaptively selects the number of events to buffer by considering both scene complexity and spatial resolution. With that said, all the methods above accumulate all events in a particular time interval without correcting for apparent motion.  
\par
A more principled approach is known as the ``motion correction''. In \cite{Gallego2018, Zhu2017}, uniform optical flow is first assumed for all events within a small spatio-temporal neighborhood. Using this assumption, an event image is then estimated by warping all events in the spatio-temporal neighborhood to a given reference time, according to the set of optical flow parameters that best describe the apparent motion of events. The optimal set of parameters is searched using iterative methods, therefore it is more computationally expensive compared to all the above methods. The uniform optical flow assumption is replaced with a spatially uniform Bézier curve motion assumption in \cite{Seok2020}. In general, a relatively large time interval (or a large number of events) is required for accurate estimation of the relative intensity change magnitude, due to inherent noise and latency of the DVS.
\par
Due to the limitations of the above event stream representation strategies, this paper adopts the concept of ``lifetime'' introduced by \cite{Mueggler2015} to represent how long will it take for an event to trigger a new associated event in its neighboring pixel location, due to persistent apparent motion. Hence, by activating pixels of the binary event image according to the estimated lifetime and position of events, a binary event image can be defined for any time instant in the past. Lifetime estimation itself can be done using the so-called ``Local Plane Fit''' \cite{Benosman2014}, which is one of the most widely adopted approaches for optical flow estimation. Normal flow is estimated by fitting a plane in a local spatio-temporal neighborhood of the Surface of Active Events (SAE), which describes the timestamp of the most recent event for each pixel location, about the incoming event. The single-shot optical flow estimation algorithm (SOFEA) \cite{Low_2020_CVPR_Workshops} considers selected events in the local spatio-temporal neighborhood with the same polarity, and thus is heavily robust to noise. 
\par
Nonetheless, the limitations of SOFEA make it rather non-ideal for direct BEI estimation. One of the main issues is that SOFEA suffers from low event density in select scenes, mainly due to refractory filtering, which implies that a significant amount of events are lost in the process. In other words, the refractory period prevents updating the SAE if an event has arrived within the refractory period. This is undesirable as the estimated BEI (similar to an edge map) is then not representative of the scene. The use of a refractory filter also limits the maximum speed of scene dynamics. Apart from that, the validity of the local planar assumption is dependent on the scene dynamics and the local scene structure. Thus, events associated to high acceleration motion or moving edges with large curvatures (e.g. a corner) will most likely be regarded as noise by the goodness-of-fit noise rejection method, as it is evaluated with respect to a planar model. Thus, we propose a pipeline for BEI estimation and lifetime augmentation, termed as SOFEA-BEI, to deal with the shortcomings of vanilla SOFEA.

\section{SOFEA-BEI}

\label{sec:bei}
An event $\boldsymbol{e}$ can be characterized by a 4-tuple $(x, y, t, p)$, where $x, y$ represents the spatial location or position of the event in pixel coordinates, $t$ represents the time of occurrence of the event and $p \in \{  -1, +1 \}$ represents the polarity of the event. On top of that, the event stream can be defined as a sequence of events $(\boldsymbol{e}_n)_{n \in \mathbb{N}}$ such that $i<j \implies t_i \leq t_j$, where $\boldsymbol{e}_i = (x_i, y_i, t_i, p_i), \boldsymbol{e}_j = (x_j, y_j, t_j, p_j)$.

\par
Disregarding event polarity, each event carries information in the three-dimensional $xyt$ spacetime domain. Consequently, the sequence of most recent events over the entire spatial domain, which is a subset of the event stream $(\boldsymbol{e}_n)_{n \in \mathbb{N}}$, can be simply described by a spatio-temporal surface $\Sigma_e$. The spatio-temporal surface $\Sigma_e$ is better known as the Surface of Active Events (SAE), as coined in \cite{Benosman2014}. The mathematical definition of the SAE is given by:
\begin{equation}
	\begin{split}
		\Sigma_e: \mathbb{R}^2 &\to\mathbb{R} \\
		\boldsymbol{p} &\mapsto \Sigma_e( \boldsymbol{p} ) = t
	\end{split}
\end{equation}
where $\boldsymbol{p} = \begin{bmatrix}x & y\end{bmatrix}^\top$.

\par
A simple expression for $\Sigma_e$ can be obtained by approximating the SAE as a plane in a local spatial neighborhood. With the approximation above, the local planar SAE can be sufficiently described by a set of plane parameters $\boldsymbol{\Pi} = \begin{bmatrix} a & b & c & d \end{bmatrix}^\top$. An interesting property of the SAE is that its spatial gradient $\nabla \Sigma_e(\boldsymbol{p}) = \begin{bmatrix} \frac{\partial \Sigma_e(\boldsymbol{p})}{\partial x} & \frac{\partial \Sigma_e(\boldsymbol{p})}{\partial y} \end{bmatrix}^\top$ encodes information of normal flow associated to the active events. For a local planar SAE, $\nabla \Sigma_e(\boldsymbol{p})$ can be further defined as $\begin{bmatrix} -\frac{a}{c} & -\frac{b}{c} \end{bmatrix}^\top$. SOFEA improved upon existing local plane fitting methods by using an efficient greedy algorithm to optimally select the spatially neighboring events associated to the incoming event for accurate non-iterative local SAE plane fitting. SOFEA also incorporated a mathematically simpler description of the local planar SAE given by the additional constraint that the incoming event lies on it. 
\par
\begin{figure}[t]
    \centering
    \includegraphics[width=0.55\linewidth]{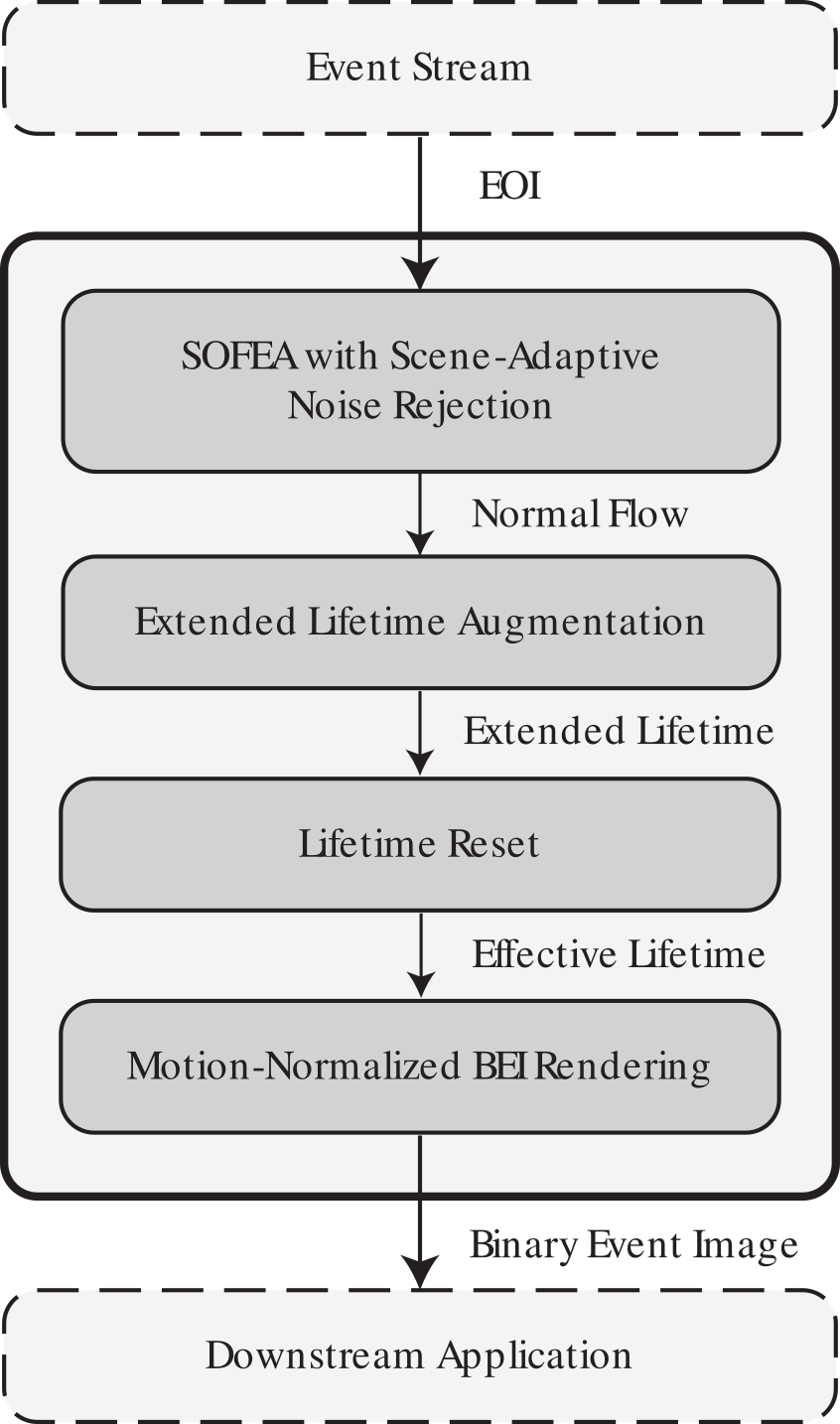}
    \caption{Pipeline of SOFEA for Binary Event Image Estimation (SOFEA-BEI).}
    \label{bei:fig:sofea_bei_pipeline}
\end{figure}

The refractory filter originally incorporated in SOFEA is removed from SOFEA-BEI so as to retain a high event density, and also to eliminate its introduced limit on the speed of scene dynamics. We then post-process the lower accuracy optical flow estimates to obtain highly satisfactory lifetimes for the purpose of BEI rendering. In particular, post-processing is performed via extended lifetime augmentation and lifetime reset. Apart from that, the goodness-of-fit noise rejection method in SOFEA is replaced with a novel scene-adaptive noise rejection method. In order to reduce visual redundancy between successive BEI renders, a motion-normalized BEI rendering time interval is also introduced. \autoref{bei:fig:sofea_bei_pipeline} illustrates the overall pipeline of SOFEA-BEI.

\subsection{Scene-adaptive Noise Rejection}
\label{bei:subsec:application}
The proposed noise rejection method for flow estimation does not depend on the goodness-of-fit criteria\cite{Aung2018} for the plane fitting, i.e. check whether the fitted plane predicts timestamps of associated events within a tolerance level. Instead, we utilize an estimated upper bound of the true underlying distribution of optical flow speed, to implicitly assess the approximate upper bound likelihood of a given optical flow speed estimate. More precisely, our method rejects events with optical flow speed estimates that fall in the tails of the approximate probability upper bound, as it is highly improbable to observe such a value. Such an approach is possible due to the fact that spurious noise events and erroneous  flow estimates generally have an extreme flow speed associated to it. Hence, this simple noise rejection method is adaptive to the scene dynamics and independent of the local scene structure with practical bounds of the true optical flow speed. 

%------------------------------------------------------------------------
\subsection{Extended Lifetime Augmentation and Reset}
\label{bei:sec:lifetime}

\par
As a variant of the Local Plane Fit method, SOFEA estimates the normal flow of an event by performing plane fitting on its corresponding local SAE. Thus, the global SAE can be approximated as a piecewise-continuous set of planes, with each plane defined for a distinct $1 \times 1$ pixel spatial neighbourhood. Theoretically, a BEI with infinite spatial resolution can then be estimated by evaluating the SAE at a particular time instant. Nevertheless, this BEI estimation approach requires highly accurate normal flow estimates, as well as high computational cost for the evaluation of each local time slice. Hence, we adopt the lifetime augmentation approach \cite{Mueggler2015}, with some refinements, for relatively accurate and efficient BEI estimation, even with low accuracy normal flow estimates.

\par
Ideally, the lifetime of an event signifies the amount of time required for an associated (new) event to be generated in its neighboring pixel location, due to apparent motion of their common corresponding feature. Suppose lifetimes of events are known, BEI can be rendered by simply activating specific pixels, given by the position of events, for a duration of their augmented lifetimes. However, event lifetimes depend on the continuous-time trajectory and velocity of the apparent motion, which are not trivial to estimate. Thus, the lifetime of an event is approximated as the maximum time required for the event to ``travel'' $1$ pixel, supposing the apparent motion velocity of the associated feature is constant, or equivalently the normal flow of the event is constant. Note that the lifetime is defined as the maximum time because in Local Plane Fit methods, only normal flow is estimated. As the normal flow  speed always underestimates the optical flow speed, only the maximum time can be estimated. Mathematically, the lifetime, $\tau$ of an event is given by:
\begin{equation}
	\tau = \frac{1}{\| \boldsymbol{v_\perp}  \|} = \| \nabla \Sigma_e(\boldsymbol{p}) \|
    \label{bei:eq:lifetime}
\end{equation}
In other words, $\tau$ is merely given by the spatial gradient magnitude of the associated local SAE.
\par
Nevertheless, the scene representation accuracy of the estimated BEI depends greatly on the accuracy of the estimated lifetimes, which in turn relies on the normal flow estimation accuracy. Apart from that, the constant velocity assumption may be violated in scenes with high acceleration or deceleration, resulting in an overestimation or underestimation of lifetimes. In order to enhance the quality of BEI estimates in general scenes and also when low accuracy normal flow estimates are used, we propose to augment each event with an extended lifetime, $\tau_\mathit{ext, bei}$ instead, given by:
\begin{equation}
	\tau_\mathit{ext, BEI} = \kappa_\mathit{ext, BEI} \ \tau, \ \kappa_\mathit{ext, BEI} \geq 1
    \label{bei:eq:ext_lifetime_bei}
\end{equation}
By doing so, it can be observed that edges in the BEI are approximately $\kappa_\mathit{ext, bei}$ pixels thick, due to the overestimation of lifetime by an approximate factor of $\kappa_\mathit{ext, BEI}$. To alleviate that, we also perform edge thinning by resetting (i.e. set to zero) the extended lifetime of the neighboring event, which is in the opposite flow direction of a given event, because the neighboring event generally indicates the ``source'' of the given event. As the optimal value of $\kappa_\mathit{ext, bei}$ might be different for accurate scene representation of the rendered BEI, and for accurate estimation of the scene complexity, given by $\widetilde{C} (t_j)$, an event can also be optionally augmented with a second extended lifetime for scene complexity estimation, as follows:
\begin{equation}
	\tau_\mathit{ext, \widetilde{C}} = \kappa_\mathit{ext, \widetilde{C}} \ \tau, \ \kappa_\mathit{ext, \widetilde{C}} \geq 1
    \label{bei:eq:ext_lifetime_C_tilde}
\end{equation}
With both lifetime extension and lifetime reset, the eventual effective lifetime, $\tau_\mathit{eff, BEI}$ or $\tau_\mathit{eff, \widetilde{C}}$ gives a relatively better estimate of the true lifetime. This is due to the larger error tolerance of normal flow estimates, given by the closed-loop nature of the lifetime estimation process. This allows for a crisp and accurate scene representation of the estimated BEI, which is visually similar to an edge map. Isolated active pixels in the estimated BEI can also be optionally filtered out, as noise in a BEI manifests as impulse noise.

\subsection{Scene-adaptive BEI Rendering}
\label{bei:sec:render}
Although \autoref{bei:sec:lifetime} allows for continuous-time BEI rendering, it is computationally infeasible and generally unnecessary, as a new BEI rendering is usually only required whenever it differs significantly from the most recent BEI rendered. Thus, a BEI rendering time interval is defined to govern the rendering time instants. Similar to how standard cameras capture intensity images at a fixed rate, it is rather reasonable to set it as a constant. Nevertheless, the amount of redundancy increases as the scene dynamics decreases. Thus, we propose a scene-adaptive BEI rendering time interval that approximately normalizes for the expected apparent displacement magnitude of features with apparent motion. In particular, it approximately normalizes for the expected number of events generated in an event substream.
%-----------------------------------------------------
\subsubsection{Event Generation Model}
\label{bei:subsec:model}
In this subsection, we will first devise a simplified model to describe how the event stream is formed by numerous features with apparent motion to assist scene-adaptive BEI rendering:
\begin{enumerate}
\item $c(t)$ is a deterministic function that defines the number of features with apparent motion (also known as the scene complexity), for all time $t$. A feature is defined as a unique surface point in the scene, within the field-of-view of the DVS, that causes events to be generated if it has apparent motion. \label{bei:def:c}
\item $\boldsymbol{v}_i(t) = \begin{bmatrix} \|\boldsymbol{v}_i(t)\| & \angle \boldsymbol{v}_i(t) \end{bmatrix}^\top$ is a deterministic vector-valued function that defines the optical flow of events associated to feature $i \in \{ \ 1, 2, \ldots, c(t)\ \}$ with apparent motion, for all time $t$, assuming infinite spatial resolution of the image sensor (hence events are generated in continuous-time). In particular, $\|\boldsymbol{v}_i(t)\|$ gives the speed and $\angle \boldsymbol{v}_i(t)$ gives the orientation of the optical flow. $\boldsymbol{v}_i(t)$ is also equivalent to the apparent motion velocity of feature $i \in \{ \ 1, 2, \ldots, c(t)\ \}$ at time $t$. \label{bei:def:v}
\item A short time interval, $\mathcal{I}(t) = ( \ t-\Delta T(t), t \ ]$ is defined for all time $t$, where $\Delta T(t)$ is a deterministic function that gives the length of the time interval. \label{bei:def:i}
\item $f_i(t)$ is a deterministic function that defines the event generation frequency associated to feature $i \in \{ \ 1, 2, \ldots, c(t)\ \}$ with apparent motion, for all time $t$. \label{bei:def:f}
\end{enumerate}
\par
With the definitions above, the assumptions are:
\begin{enumerate}
\item $c(t')$ and $\boldsymbol{v}_i(t'), \forall i \in \{ \ 1, 2, \ldots, c(t')\ \}$ have the same respective values for all time $t'$ in $\mathcal{I}(t)$. In other words, $c(t)$ and $\boldsymbol{v}_i(t), \forall i \in \{ \ 1, 2, \ldots, c(t)\ \}$ are constant throughout $\mathcal{I}(t)$. \label{bei:assmp:constant_c_v}
\item Given that $\boldsymbol{v}_i(t')$ has the same value for all time $t'$ in $\mathcal{I}(t)$, $f_i(t')$ also has the same value. In other words, if $\boldsymbol{v}_i(t)$ is constant throughout $\mathcal{I}(t)$, it is also the case for $f_i(t)$. Implicitly, this assumes that all pixels have the same contrast threshold, which could be negligibly violated in real DVS data due to random hardware mismatch. This assumption also disregards the subtle variation of $f_i(t')$ if the apparent motion direction of feature $i$, or equivalently $\angle \boldsymbol{v}_i(t')$, is non-Ordinal or non-Cardinal with respect to the DVS pixel grid. \label{bei:assmp:constant_f} 
\end{enumerate}
Assumptions \ref{bei:assmp:constant_c_v} and \ref{bei:assmp:constant_f} can be combined to yield the following: $c(t'), \boldsymbol{v}_i(t')$ and $f_i(t')$, $\forall i \in \{ \ 1, 2, \ldots, c(t')\ \}$ have the same respective values for all time $t'$ in $\mathcal{I}(t)$. As a result, the event generation frequency of the event stream, given by:
\begin{equation}
	f(t') = \sum\limits_{i=1}^{c(t')} f_i(t')
    \label{bei:eq:f}
\end{equation}
also has the same value for all time $t'$ in $\mathcal{I}(t)$. Also, it can be observed that a direct relationship exists between $\|\boldsymbol{v}_i(t)\|$ and $f_i(t)$. In fact, it is given by:
\begin{equation}
	\|\boldsymbol{v}_i(t)\| = \alpha(\angle \boldsymbol{v}_i(t)) \ f_i(t)
    \label{bei:eq:v_f_relation}
\end{equation}
where $\alpha(\theta)$ is a deterministic function that specifies the conversion from frequency to pixel units depending on the orientation of optical flow.
%------------------------------------------------------
\subsection{Scene-adaptive Rendering}
Scene-adaptive motion normalization can be achieved between successive BEI renders by using the expected number of events generated in an event substream within a small temporal window. To formally derive the scene-adaptive BEI rendering time interval, we first present the following definitions:
\begin{enumerate}
\item A short time interval, $\mathcal{I}_\mathit{BEI} (t) = [ \ t, \ t + \Delta T_\mathit{BEI} (t) \ )$ is defined for all time $t$, where $\Delta T_\mathit{BEI} (t)$ is a deterministic function that gives the length of the time interval (the scene-adaptive BEI rendering time interval).
\item $\mathbf{s}_i(t)$ is a deterministic vector-valued function that defines the apparent displacement of feature $i \in \{ \ 1, 2, \ldots, c(t)\ \}$ with apparent motion from the start to the end of the time interval $\mathcal{I}_\mathit{BEI} (t)$ (i.e. between time $t$ and $t + \Delta T_\mathit{BEI} (t)$). Mathematically, $\mathbf{s}_i(t)$ is given by:
\begin{equation}
	\mathbf{s}_i(t) = \int_t^{t + \Delta T_\mathit{BEI} (t)}
    	\begin{bmatrix}
			\boldsymbol{v}_{i, x}(t') \\
            		  \boldsymbol{v}_{i, y}(t')
		\end{bmatrix}
        	  dt'
              \label{bei:eq:s}
\end{equation}
\item $\{ \ S(t),\ t \in [ \ 0, \infty \ ) \ \}$ is a continuous stochastic process, whereby it is equally likely for $S(t)$ to give any of $\|\boldsymbol{s}_i(t)\|, \ i \in \{ \ 1, 2, \ldots, c(t)\ \}$. In other words, $S(t)$ describes the distribution of $\|\boldsymbol{s}_i(t)\|$. With that, the  expected value of $S(t)$ is given by: \label{bei:def:S}
\begin{equation}
	\mathbb{E} [ S(t) ] = \frac{1}{c(t)} \sum\limits_{i=1}^{c(t)} \|\boldsymbol{s}_i(t)\|
	\label{bei:eq:exp_S}
\end{equation}

\item $\{ \ F(t),\ t \in [ \ 0, \infty \ ) \ \}$ is a continuous stochastic process, whereby it is equally likely for $F(t)$ to give any of $f_i(t), \ i \in \{ \ 1, 2, \ldots, c(t)\ \}$. In other words, $F(t)$ describes the distribution of $f_i(t)$. Thus, the expected value of $F(t)$ is given by (\autoref{bei:eq:f}):
\begin{equation}
	\mathbb{E} [ F(t) ] = \frac{1}{c(t)} \sum\limits_{i=1}^{c(t)} f_i(t) = \frac{f(t)}{c(t)}
    		  \label{bei:eq:exp_F}
\end{equation}
\label{bei:def:F}
\item $\{ \ V(t),\ t \in [ \ 0, \infty \ ) \ \}$ is a continuous stochastic process, whereby it is equally likely for $V(t)$ to give any of $\|\boldsymbol{v}_i(t)\|, \ i \in \{ \ 1, 2, \ldots, c(t)\ \}$. In other words, $V(t)$ describes the distribution of $\|\boldsymbol{v}_i(t)\|$. With that, the  expected value of $V(t)$ is given by (\autoref{bei:eq:v_f_relation}):
\begin{equation}
	\mathbb{E} [ V(t) ] = \frac{1}{c(t)} \sum\limits_{i=1}^{c(t)} \|\boldsymbol{v}_i(t)\| = \frac{1}{c(t)} \sum\limits_{i=1}^{c(t)} \alpha(\angle \boldsymbol{v}_i(t)) \ f_i(t)
	\label{bei:eq:exp_V}
\end{equation}
Because $\alpha(\angle \boldsymbol{v}_i(t)) \in [ \ 1, \sqrt{2} \ ]$, the following inequality regarding $\mathbb{E} [ V(t) ]$ and $\mathbb{E} [ F(t) ]$ holds:
\begin{equation}
	\mathbb{E} [ F(t) ] \leq \mathbb{E} [ V(t) ] \leq \sqrt{2} \ \mathbb{E} [ F(t) ]
	\label{bei:eq:exp_v_exp_f_ineq}
\end{equation}
\end{enumerate}
Along with the above definitions, we impose similar assumptions, as given by Assumptions \ref{bei:assmp:constant_c_v} and \ref{bei:assmp:constant_f} in \autoref{bei:subsec:model}, but on the time interval $\mathcal{I}_\mathit{BEI} (t)$ instead. Thus, \autoref{bei:eq:s} can be simplified to:
\begin{equation}
	\mathbf{s}_i(t) = 
    	\begin{bmatrix}
			\boldsymbol{v}_{i, x}(t') \\
            		  \boldsymbol{v}_{i, y}(t')
		\end{bmatrix}
        	  \Delta T_\mathit{BEI} (t)
              \label{bei:eq:s_simple}
\end{equation}
Besides, \autoref{bei:eq:exp_S} can also be simplified to give:
\begin{equation}
	\mathbb{E} [ S(t) ] = \frac{1}{c(t)} \sum\limits_{i=1}^{c(t)} \|\boldsymbol{v}_i(t)\| \ \Delta T_\mathit{BEI} (t)
	\label{bei:eq:exp_S_simple} = \mathbb{E} [ V(t) ] \ \Delta T_\mathit{BEI} (t)
\end{equation}
Substituting \autoref{bei:eq:exp_v_exp_f_ineq} into the above equation, the following bounds can be imposed on $\mathbb{E} [ S(t) ]$:
\begin{equation}
	\mathbb{E} [ F(t) ] \ \Delta T_\mathit{BEI} (t) \leq \mathbb{E} [ S(t) ] \leq \sqrt{2} \ \mathbb{E} [ F(t) ] \ \Delta T_\mathit{BEI} (t)
	\label{bei:eq:exp_s_exp_f_ineq}
\end{equation}
Note that $\mathbb{E} [ F(t) ] \ \Delta T_\mathit{BEI} (t)$ actually gives the expected number of events generated in an event substream within $\mathcal{I}_\mathit{BEI} (t)$ (proof omitted). Similar to the case of $\Delta T (t)$, the choice of $\Delta T(t)_\mathit{BEI}$ is arbitrary, despite being deterministic. Therefore, we suppose that $\Delta T(t)_\mathit{BEI}$ is defined such that it coincides with:
\begin{equation}
	\Delta T(t)_\mathit{BEI} = \frac{2}{1 + \sqrt{2}} \cdot \frac{ \overline{s} }{ \hat{F} (t) }
    \label{bei:eq:delta_t_bei}
\end{equation}
in the first realization of $\hat{F} (t)$, where $\overline{s}$ is a constant representing the desired value of $\mathbb{E} [ S(t) ]$. Substituting \autoref{bei:eq:delta_t_bei} into \autoref{bei:eq:exp_s_exp_f_ineq} gives the following:
\begin{equation}
	\frac{2}{1 + \sqrt{2}} \cdot \frac{ \mathbb{E} [ F(t) ] }{ \hat{F} (t) } \cdot \overline{s} \leq \mathbb{E} [ S(t) ] \leq \frac{2  \sqrt{2}}{1 + \sqrt{2}} \cdot \frac{ \mathbb{E} [ F(t) ] }{ \hat{F} (t) } \cdot \overline{s}
	\label{bei:eq:exp_s_exp_f_ineq_expanded}
\end{equation}
Thus, $\Delta T(t)_\mathit{BEI}$ is defined in such a way that $\overline{s}$ is the midpoint of the above bounds on $\mathbb{E} [ S(t) ]$. Equivalently, it is also defined such that $\frac{2}{1 + \sqrt{2}} \overline{s}$ is the desired expected number of events generated in an event substream within $\mathcal{I}_\mathit{BEI} (t)$ (i.e. suppose the mean squared error of $\hat{F}(t)$ is kept small relative to $\mathbb{E} [ F(t) ]$, $\mathbb{E} [ F(t) ] \ \Delta T_\mathit{BEI} (t) \approx \frac{2}{1 + \sqrt{2}} \overline{s}$). Therefore, motion normalization can be achieved between successive BEI renders by using $\Delta T(t)_\mathit{BEI}$ as the rendering interval. 

%---------------------------------------------------------------
\begin{figure}[t]
    \centering
		\includegraphics[width=1\linewidth]{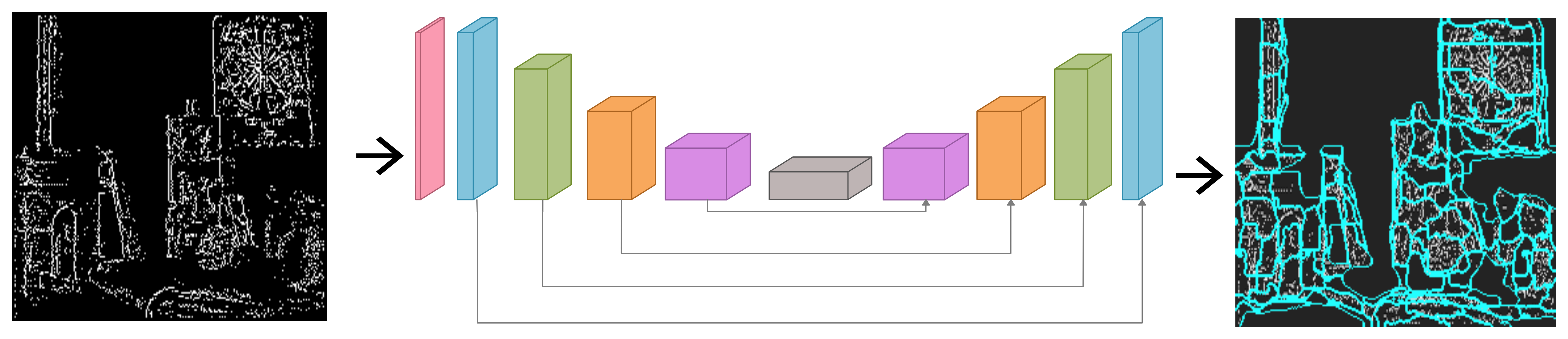} 
    \caption{Schematic representation of the FCN used for extracting superevents from SOFEA-BEI}
    \label{fig:network_fig}
\end{figure}
%--------------------------------
\section{SOFEA-BEI-FCN Superevents}
\label{sec:superevents}
In this section, we introduce our CNN-based superevents segmentation method using the output of SOFEA-BEI. We make use of the architecture proposed in \cite{yang2020sup}, which has been shown to perform well on RGB compared to several state-of-the-art methods. Superpixels as defined for standard images are supposed to delineate the edges, and thus binary images containing edge-like structures can also be well-segmented by the fully convolutional neural network architecture proposed in \cite{yang2020sup}.
\par
The model in \cite{yang2020sup} divides a given image into a number of grids with preset dimensions. These grids form the initial superpixels. The model then makes use of a series of convolutions followed by one of deconvolutions in order to achieve the desired segmentation, as illustrated in Figure \ref{fig:network_fig}. Further, features from earlier layers are concatenated to those from deeper layers via skip connections. The final output is an assignment map 

\begin{equation}
	\begin{split}
		g: \mathbb{Z}^{H \times W} &\to N_r \\
		\boldsymbol{r} &\mapsto g( \boldsymbol{r} ) = n
	\end{split}
\end{equation}
where $\boldsymbol{r} = (y, x)$ and $n$ denotes the superpixel that each pixel has been assigned to. Also, $N_r$ denotes the nine superpixels in the neighbourhood of the given pixel. A lot of the simplicity in \cite{yang2020sup} is the direct result of this part of their formulation: any given pixel can only be assigned to one of the nine superpixels in its immediate neighbourhood. This results in compact superpixels and oversegmenation to an extent that is greater than what is observed in traditional superpixel segmentation methods. \cite{yang2020sup} justify this by arguing that since the superpixels thus obtained will eventually be used in downstream tasks, the latter may group perceptually similar superpixels if such a need arises. Hence, this formulation does not compromise on the quality of superpixels for use in downstream tasks, while also helping to decrease the number of training and inference time computations.
\par
For the sake of better visualization in Figure \ref{fig:network_fig}, superevents with little to no events have been merged. In practice, this was achieved by computing the ratio of the number of active events to that of the total number of pixels inside a given superevent. They were merged together whenever this ratio turned out to be less than a pre-determined threshold (set to 0.05 in this work). 

\begin{figure}[t]
	\centering 
    
	\begin{subfigure}{0.38\linewidth}
    	\centering
		\includegraphics[width=1\linewidth]{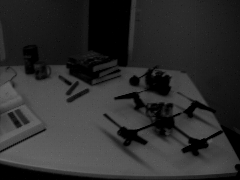} 
		\caption{APS intensity image}
		\label{bei:fig:dynamic_6dof_aps}
	\end{subfigure} \hspace{1em}
	\begin{subfigure}{0.38\linewidth}
    	\centering
        \includegraphics[width=1\linewidth]{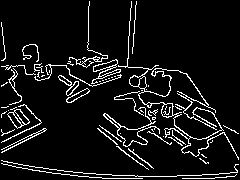}
		\caption{Edge map via \cite{Canny1986}}
		\label{bei:fig:dynamic_6dof_canny}
	\end{subfigure}

    \begin{subfigure}{0.38\linewidth}
    	\centering
		\includegraphics[width=1\linewidth]{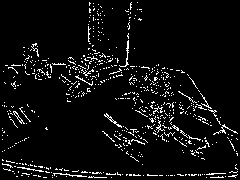} 
		\caption{$\kappa_\mathit{ext, BEI} = 1$, no reset}
		\label{bei:fig:dynamic_6dof_sofea_bei_ext1_noreset}
	\end{subfigure} \hspace{1em}
	\begin{subfigure}{0.38\linewidth}
    	\centering
        \includegraphics[width=1\linewidth]{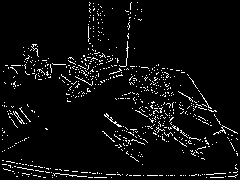}
		\caption{$\kappa_\mathit{ext, BEI} = 1$, with reset}
		\label{bei:fig:dynamic_6dof_sofea_bei_ext1_reset}
	\end{subfigure}
    
	\begin{subfigure}{0.38\linewidth}
    	\centering
		\includegraphics[width=1\linewidth]{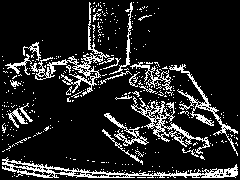} 
		\caption{$\kappa_\mathit{ext, BEI} = 3$, no reset}
		\label{bei:fig:dynamic_6dof_sofea_bei_ext3_noreset}
	\end{subfigure} \hspace{1em}
	\begin{subfigure}{0.38\linewidth}
    	\centering
        \includegraphics[width=1\linewidth]{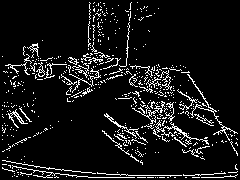}
		\caption{$\kappa_\mathit{ext, BEI} = 3$, with reset}
		\label{bei:fig:dynamic_6dof_sofea_bei_ext3_reset}
	\end{subfigure}
    
	\begin{subfigure}[t]{0.38\linewidth}
    	\centering
        \includegraphics[width=1\linewidth]{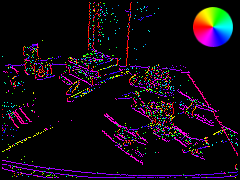}
		\caption{$\kappa_\mathit{ext, BEI} = 3$, with reset, augmented with normal flow orientation}
		\label{bei:fig:dynamic_6dof_sofea_bei_flow_ext3_reset}
	\end{subfigure} \hspace{1em}  
	\begin{subfigure}[t]{0.38\linewidth}
    	\centering
		\includegraphics[width=1\linewidth]{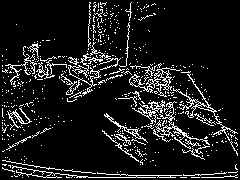} 
		\caption{$\kappa_\mathit{ext, BEI} = 5$, with reset}
		\label{bei:fig:dynamic_6dof_sofea_bei_ext5_reset}
	\end{subfigure}
 
	\caption{Comparison of BEI estimated using SOFEA-BEI, under various extended lifetime augmentation and lifetime reset configurations, on the {\pcrFont dynamic\_6dof} sequence.}
	\label{bei:fig:lifetime_analysis}
\end{figure}

\begin{figure}[t]
	\centering 
    
    \begin{minipage}[t]{0.29\linewidth}
		\begin{subfigure}{1\linewidth}
    		\centering
			\includegraphics[width=1\linewidth]{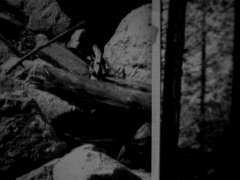} 
			\caption{}
			\label{bei:fig:slider_far_aps}
		\end{subfigure}
    
		\begin{subfigure}{1\linewidth}
            \centering
            \includegraphics[width=1\linewidth]{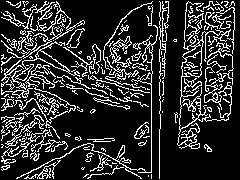}
            \caption{}
            \label{bei:fig:slider_far_canny}
		\end{subfigure}
        
		\begin{subfigure}{1\linewidth}
            \centering
            \includegraphics[width=1\linewidth]{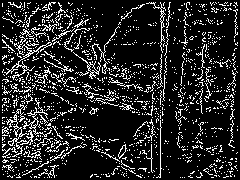}
            \caption{}
            \label{bei:fig:slider_far_sofea_bei}
		\end{subfigure}
        
	%	\begin{subfigure}{1\linewidth}
    %        \centering
   %         \includegraphics[width=1\linewidth]{{figures/sofea_bei_filt_slider_far_70}}
  %          \caption{}
 %           \label{bei:fig:slider_far_sofea_bei_filt}
%		\end{subfigure}
        
		\begin{subfigure}{1\linewidth}
            \centering
            \includegraphics[width=1\linewidth]{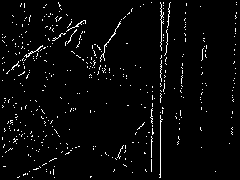}
            \caption{}
            \label{bei:fig:slider_far_mueggler2015}
		\end{subfigure}
    \end{minipage} \hspace{1em}
    \begin{minipage}[t]{0.29\linewidth}
		\begin{subfigure}{1\linewidth}
    		\centering
			\includegraphics[width=1\linewidth]{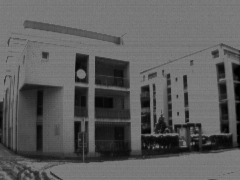} 
			\caption{}
			\label{bei:fig:urban_aps}
		\end{subfigure}
    
		\begin{subfigure}{1\linewidth}
            \centering
            \includegraphics[width=1\linewidth]{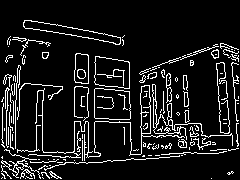}
            \caption{}
            \label{bei:fig:urban_canny}
		\end{subfigure}
        
		\begin{subfigure}{1\linewidth}
            \centering
            \includegraphics[width=1\linewidth]{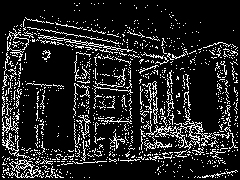}
            \caption{}
            \label{bei:fig:urban_sofea_bei}
		\end{subfigure}
        
%		\begin{subfigure}{1\linewidth}
%            \centering
%            \includegraphics[width=1\linewidth]{{figures/sofea_bei_filt_urban_100_bei}}
%            \caption{}
%            \label{bei:fig:urban_sofea_bei_filt}
%		\end{subfigure}
        
		\begin{subfigure}{1\linewidth}
            \centering
            \includegraphics[width=1\linewidth]{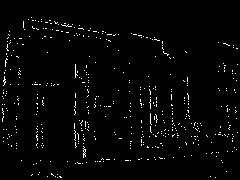}
            \caption{}
            \label{bei:fig:urban_mueggler2015}
		\end{subfigure}
    \end{minipage} \hspace{1em}
    \begin{minipage}[t]{0.29\linewidth}
		\begin{subfigure}{1\linewidth}
    		\centering
			\includegraphics[width=1\linewidth]{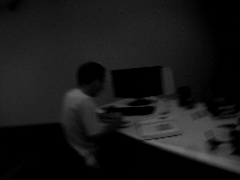} 
			\caption{}
			\label{bei:fig:dynamic_rotation_aps}
		\end{subfigure}
    
		\begin{subfigure}{1\linewidth}
            \centering
            \includegraphics[width=1\linewidth]{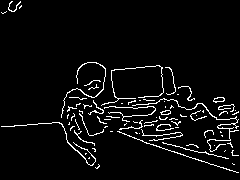}
            \caption{}
            \label{bei:fig:dynamic_rotation_canny}
		\end{subfigure}
        
		\begin{subfigure}{1\linewidth}
            \centering
            \includegraphics[width=1\linewidth]{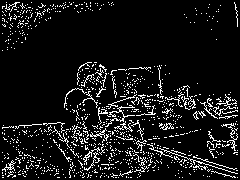}
            \caption{}
            \label{bei:fig:dynamic_rotation_sofea_bei}
		\end{subfigure}
        
%		\begin{subfigure}{1\linewidth}
%            \centering
%            \includegraphics[width=1\linewidth]{{figures/sofea_bei_filt_dynamic_rotation_1000}}
%            \caption{}
%            \label{bei:fig:dynamic_rotation_sofea_bei_filt}
%		\end{subfigure}
        
		\begin{subfigure}{1\linewidth}
            \centering
            \includegraphics[width=1\linewidth]{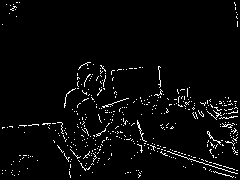}
            \caption{}
            \label{bei:fig:dynamic_rotation_mueggler2015}
		\end{subfigure}
    \end{minipage}    
    
	\caption[BEI scene representation accuracy comparison between SOFEA-BEI and \cite{Mueggler2015} on the {\pcrFont slider\_far}, {\pcrFont urban} and {\pcrFont dynamic\_rotation} sequences.]{BEI scene representation accuracy comparison between SOFEA-BEI (3\textsuperscript{rd} row) and \cite{Mueggler2015} (4\textsuperscript{th} row) on the {\pcrFont slider\_far} (1\textsuperscript{st} column), {\pcrFont urban} (2\textsuperscript{nd} column) and {\pcrFont dynamic\_rotation} (3\textsuperscript{rd} column) sequences. APS intensity images (1\textsuperscript{st} row) and edge maps obtained via Canny edge detection (2\textsuperscript{nd} row) are also visualized.}
	\label{bei:fig:bei_comparison_2}
\end{figure}
\section{Experiments}

\label{sec:experiments}
We use sequences from the event camera dataset \cite{Mueggler2017} for testing SOFEA-BEI and subsequently the superevents extraction using the FCN. The superevents concept is introduced for the first time using event cameras, to the best of our knowledge. Thus, evaluation is focused on comparing SOFEA-BEI-FCN to SOFEA-BEI-SLIC \cite{slicachanta2012} and APS-FCN with simultaneously recorded intensity frames from the publicly available dataset (upper baseline). The ground truth segmentation results for SOFEA-BEI are obtained from the corresponding APS-Canny edge maps for the calculation of boundary precision and recall \cite{yang2020sup} with the natural assumption that APS frames have more information for segmentation compared to BEI. Next we qualitatively evaluate SOFEA-BEI results comapared to vanilla SOFEA.

\begin{figure*}[t]
	\centering 
    
	\begin{subfigure}{0.3\linewidth}
    	\centering
		\includegraphics[width=1\linewidth]{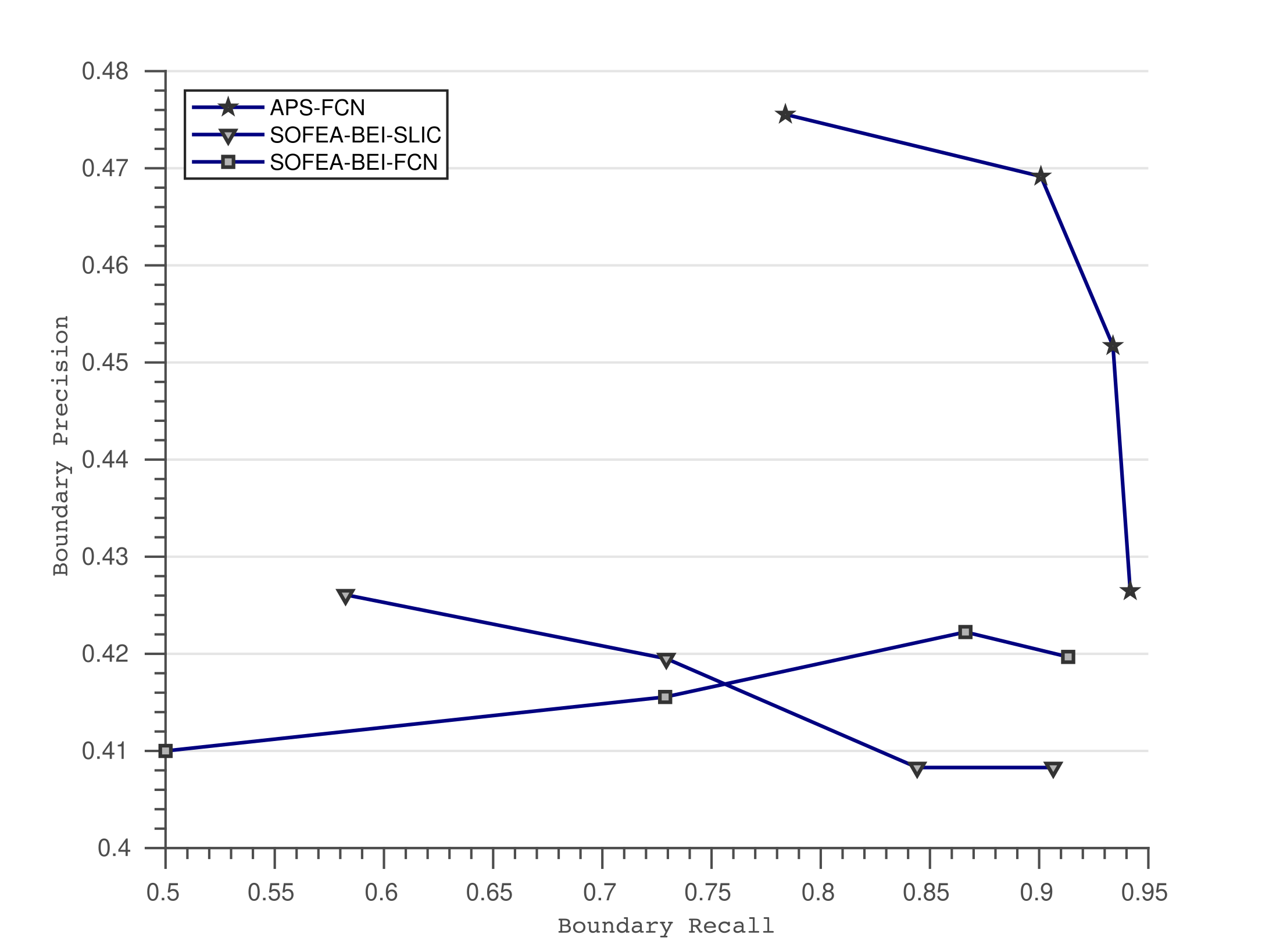} 
		\caption{{\pcrFont slider\_depth}}
		\label{bei:fig:slider_depth}
	\end{subfigure} 
		\begin{subfigure}{0.31\linewidth}
    	\centering
		\includegraphics[width=1\linewidth]{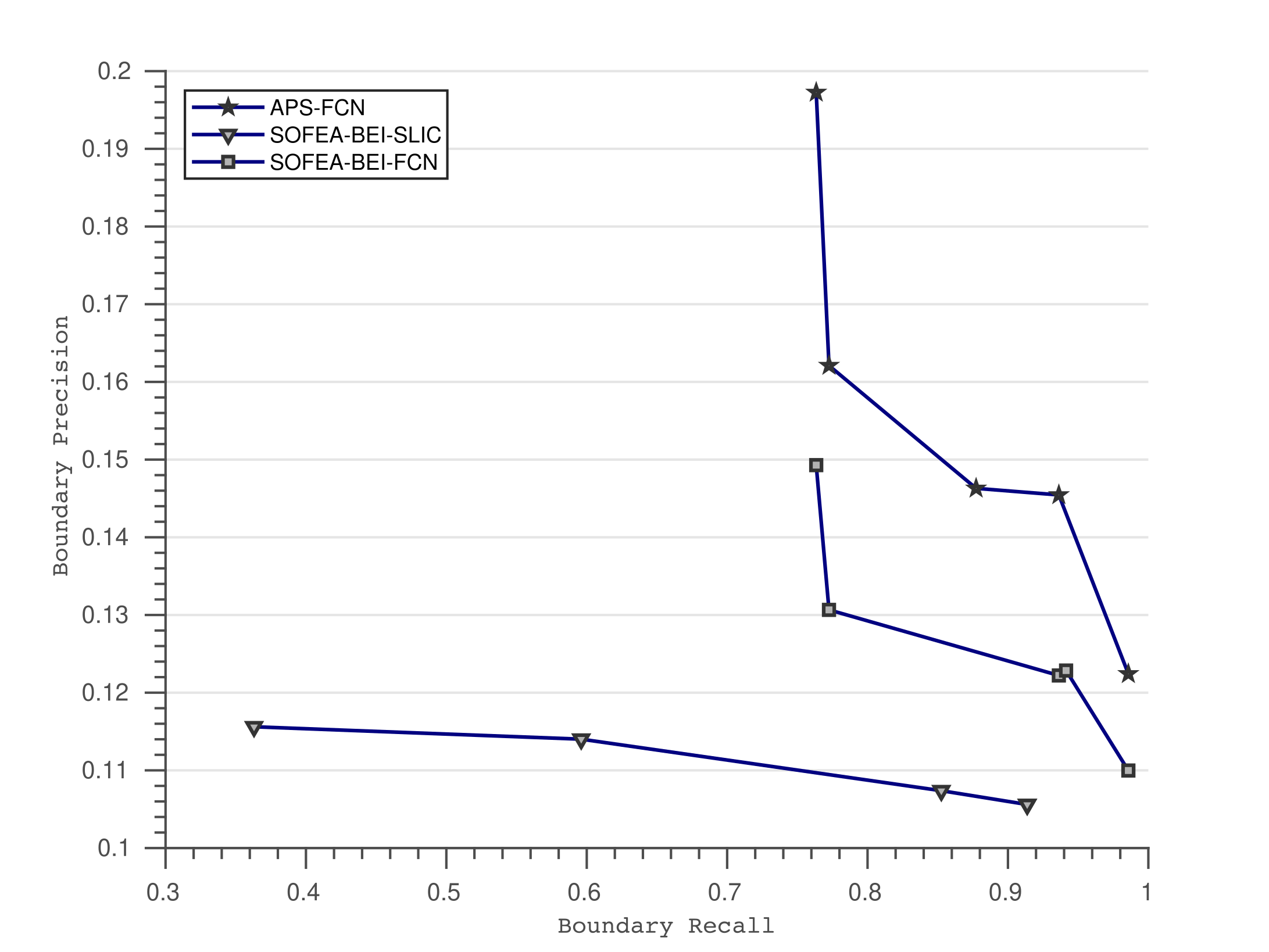} 
		\caption{{\pcrFont shapes\_translation}}
		\label{bei:fig:slider_trans}
	\end{subfigure} 
   \begin{subfigure}{0.33\linewidth}
    	\centering
		\includegraphics[width=1\linewidth]{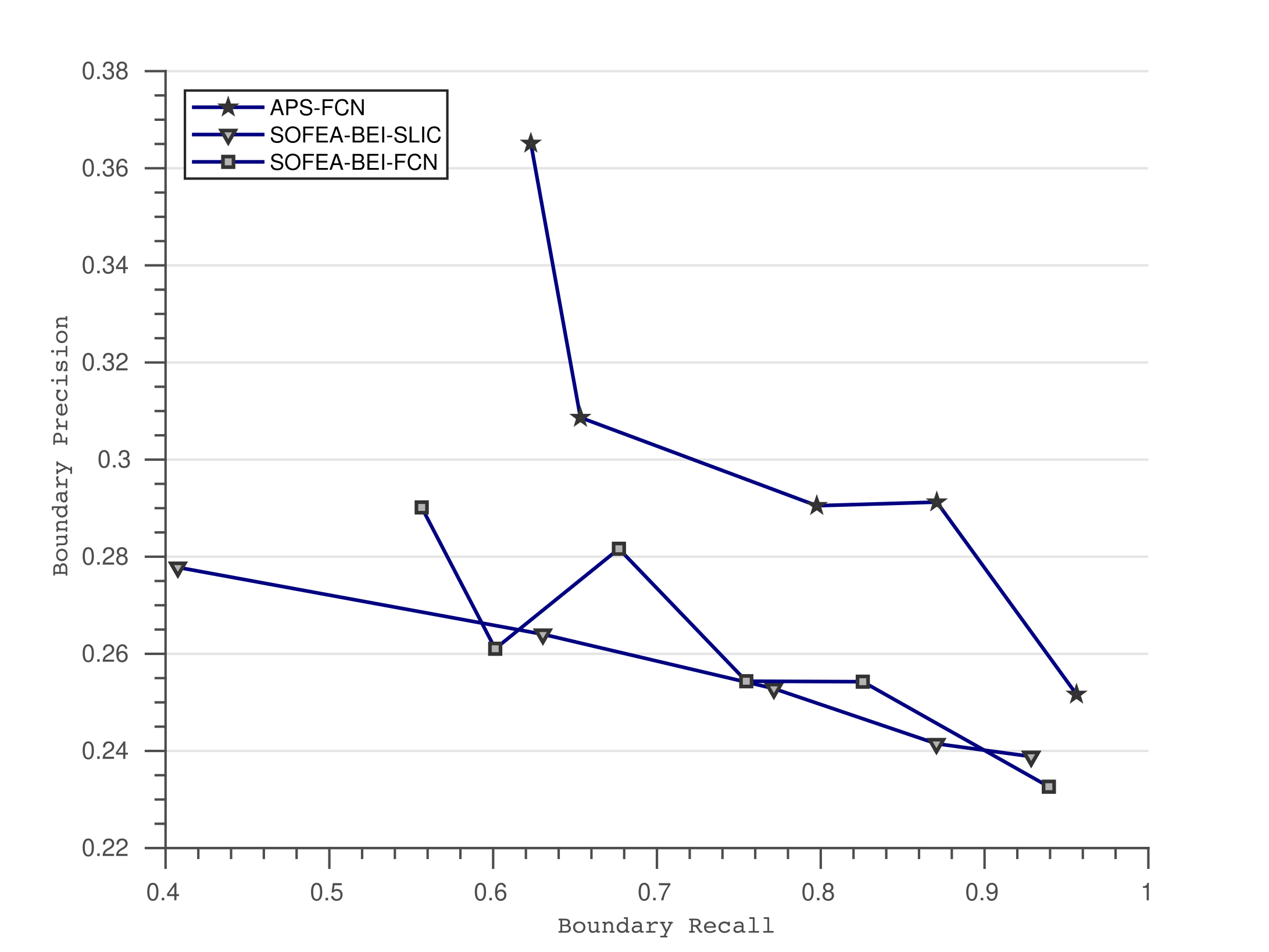} 
		\caption{{\pcrFont dynamic\_rotation}}
		\label{bei:fig:dyn_rot}
	\end{subfigure} 
		\caption{Comparison of SOFEA-BEI superevents, using FCN and SLIC methods, to APS-FCN superpixels.}
	\label{bei:fig:superevents_analysis}
\end{figure*}
\subsection{Extended Lifetime Augmentation \& Reset}
\label{bei:subsec:exp_ext_lifetime_augm_reset}
We carry out a qualitative analysis on the effectiveness of extended lifetime augmentation and reset in SOFEA-BEI for enhanced BEI estimation accuracy via the {\pcrFont dynamic\_6dof} sequence \cite{Mueggler2017}. \autoref{bei:fig:lifetime_analysis} illustrates the BEI estimated using SOFEA-BEI, under various extended lifetime augmentation and lifetime reset configurations, on the {\pcrFont dynamic\_6dof} sequence. An APS intensity image of the scene is shown in \autoref{bei:fig:dynamic_6dof_aps}, and its corresponding edge map generated using Canny edge detection is given in \autoref{bei:fig:dynamic_6dof_canny}. Since an edge map of the APS image gives a good representation of a BEI, we consider \autoref{bei:fig:dynamic_6dof_canny} as the reference for the BEI estimates.  It is worth noting that there is no fixed ``sampling interval'' for the whole pipeline, as SOFEA-BEI is an event-based method adapted to dynamically handle object/camera motion. 
%With that said, there might be details captured in a BEI but not in an edge map as well, due to limitations of the edge detection, motion blur in the APS image, high dynamic range of the scene and other reasons. 
\par
\autoref{bei:fig:dynamic_6dof_sofea_bei_ext1_noreset} illustrates the BEI estimated using the original lifetime augmentation approach \cite{Mueggler2015} (equivalently, without lifetime extension or $\kappa_\mathit{ext, BEI} = 1$) without performing lifetime reset, while \autoref{bei:fig:dynamic_6dof_sofea_bei_ext1_reset} shows the BEI estimated with lifetime reset done  for edge thinning. It can be observed that edges in both BEI estimates are generally one pixel thick, as opposed to BEI estimated using the direct accumulation approach (\autoref{fig:intro_fig}). However, several pixel thick edges still exist in \autoref{bei:fig:dynamic_6dof_sofea_bei_ext1_noreset} due to the overestimation of lifetime. This issue is almost completely rectified after lifetime reset is incorporated, which produces a more crisp representation of the scene. 
\par
On the other hand, extended lifetime augmentation is performed in \autoref{bei:fig:dynamic_6dof_sofea_bei_ext3_reset} to \autoref{bei:fig:dynamic_6dof_sofea_bei_ext5_reset}. Observe that edges in \autoref{bei:fig:dynamic_6dof_sofea_bei_ext3_noreset} are approximately three pixels thick with $\kappa_\mathit{ext, BEI} = 3$ and lifetime reset not done. The incorporation of lifetime reset in \autoref{bei:fig:dynamic_6dof_sofea_bei_ext3_reset} introduces a substantial improvement in terms of scene representation accuracy, when compared to \autoref{bei:fig:dynamic_6dof_sofea_bei_ext1_reset}. Note that the amount of detail captured is now similar to that of the edge map. Nevertheless, the amount of noise increases as well, due to events with extended lifetimes not being reset. Fortunately, there is still an apparent increase in the Signal-to-Noise Ratio (SNR). \autoref{bei:fig:dynamic_6dof_sofea_bei_flow_ext3_reset} augments \autoref{bei:fig:dynamic_6dof_sofea_bei_ext3_reset} with normal flow orientation estimates, given by the color of the active pixels according to the color wheel displayed. Observe that active pixels associated to the same edge have a common color, which corresponds to the direction perpendicular to the edge. This is desirable for an effective lifetime reset. \autoref{bei:fig:dynamic_6dof_sofea_bei_ext5_reset} illustrates the BEI estimated with $\kappa_\mathit{ext, BEI} = 5$ and lifetime reset. Note that the amount of detail captured is almost the same as that of $\kappa_\mathit{ext, BEI} = 3$, despite a significant increase of noise until the point where feature tracks can be clearly seen. This suggests that $\kappa_\mathit{ext, BEI} = 3$ and lifetime reset is sufficient for SOFEA-BEI to achieve accurate BEI estimation.
\par
To validate SOFEA-BEI, \cite{Mueggler2015} is employed as the benchmark to evaluate its performance. In addition, edge maps generated using Canny edge detection on the APS intensity image also serves as a ground truth reference for BEI estimates. The qualitative comparison is performed with \cite{Mueggler2015} on three sequences from \cite{Mueggler2017}. This collection of sequences covers a wide range of scenes with varying complexity, dynamic range and motion profiles. The details captured in a BEI need not be captured in the edge map, due to limitations of the edge detection, motion blur in the APS image, and high dynamic range of the scene.
\par
\autoref{bei:fig:bei_comparison_2} illustrates the comparison on the three mentioned sequences. In general, it can be observed that SOFEA-BEI estimates achieve remarkable scene representation, similar to the corresponding edge maps. BEI estimates of \cite{Mueggler2015} however are generally sparse in nature, as it fails to capture the abundance of detail in the scene. This is heavily reflected in \autoref{bei:fig:slider_far_mueggler2015} and \autoref{bei:fig:urban_mueggler2015}. As noted earlier, there exist details not captured in the edge map, but reflected in the SOFEA-BEI estimates. For instance, the concentric rings of the dartboard in the {\pcrFont slider\_depth} sequence, the top left edges of the building in the {\pcrFont urban} sequence and the details of objects in the {\pcrFont dynamic\_rotation} sequence. This is made possible due to the high dynamic range of the event stream and accurate SOFEA-BEI estimates.
%--------------------------------------------------------------------------------

\subsection{Superevents}
\label{subsec:supeventsresults}
\autoref{bei:fig:superevents_analysis} illustrates the boundary precision and recall for different sequences, which measures the boundary adherence of the superevents/pixels according to the ground truth with a 3$\times$3 tolerance window as done in earlier works \cite{yang2020sup}. The performance of the APS frames using FCN is much better compared to its BEI counterpart using the event stream, owing to the richness of the intensity information. This does not imply that the event stream is lacking information, but rather the way it is used in this work as a BEI. To investigate whether the lower performance is due to the use of the pre-trained network from \cite{yang2020sup}, the popular hand-crafted SLIC \cite{slicachanta2012} was used on BEI to extract superevents. This doubt is ruled out as SOFEA-BEI-SLIC has generally lower performance, especially in the shapes dataset where the edge maps form a strong ground truth. A short clip of the superevents can be viewed here\footnote{\url{https://tinyurl.com/hv3bkz5h}}. This video shows good boundary separation between the superevents at different depths, especially when a new object closer to the camera enters the field-of-view. 
\par
To the best of our knowledge, we have introduced the first mid-level perceptual grouping of events using event-based flow estimation that does not limit the rate at which superevents are extracted. In theory, superevents can be extracted for every flow estimate using the proposed scheme although the scene-adaptive BEI is better suited for generating superevents regardless of scene and camera dynamics. Overall, superevents have several potential applications, including native semantic segmentation, generating templates for object tracking, depth estimation, feature extraction, etc., and showcasing them is beyond the scope of this paper. 

\section{Conclusion}
\label{sec:conc}
This paper proposed the first superevents extraction method for event cameras, leveraging on a robust binary event image generation scheme and fully convolutional networks. This opens up a wide range of possibilities in the neuromorphic vision domain to aggregate sparse, disjoint visual information into more meaningful intermediate representations. We evaluated the superevents on the event camera dataset and highlighted the gap in boundary scores between the intensity and events modalities. Future work to address this gap using completely asynchronous processing shall be explored.

%\begin{acks}
%To Robert, for the bagels and explaining CMYK and color spaces.
%\end{acks}

\bibliographystyle{ACM-Reference-Format}
\bibliography{sample-base}


\end{document}